%% file: main.tex
\documentclass{bmvc2k}

\usepackage{chngcntr}
\usepackage{multirow}
\usepackage{booktabs}
\usepackage{siunitx}
\usepackage{makecell}
\usepackage{amssymb}
\usepackage{pifont}
\newcommand{\envelope}{\ding{41}}
\makeatletter\long\def\addinstitutionmark#1#2{\advance\bmv@ninsts1 \bmv@savenewbox{inst\the\bmv@ninsts}{\bmv@RenderInst{#2}}\bmv@savenewbox{instFN\the\bmv@ninsts}{\sffamily\ensuremath{\rule{0pt}{1ex}^{\mbox{\scriptsize #1}}\,}\hbox{\bmv@RenderInst{#2}}}\bmv@edefappend{\bmv@insts}{{\the\bmv@ninsts}}}\makeatother
\usepackage[section]{placeins}
\newcommand{\bftab}{\fontseries{b}\selectfont}

\title{Conditional Diffusion for 3D CT Volume Reconstruction from 2D X-rays}

\addauthor{Martin Rath}{martin.rath@tum.de}{1,2,*,{\envelope}}
\addauthor{Morteza Ghahremani}{morteza.ghahremani@tum.de}{1,2,*}
\addauthor{Yitong Li}{yi_tong.li@tum.de}{1,2,*}
\addauthor{Ashkan Taghipour}{ashkan.taghipour@research.uwa.edu.au}{3}
\addauthor{Marcus Makowski}{marcus.makowski@tum.de}{1}
\addauthor{Christian Wachinger}{christian.wachinger@tum.de}{1,2}

\addinstitution{
 Technical University \\
 of Munich (TUM)\\
 Munich, Germany
}
\addinstitution{
Munich Center for \\
Machine Learning (MCML)\\
Munich, Germany
}
\addinstitution{
University of Western \\
Australia (UWA)\\
Perth, Australia
}
\addinstitutionmark{*}{Equal Contribution.}
\addinstitutionmark{\envelope}{Corresponding author}

\runninghead{Rath et al}{3D CT Volume Reconstruction from 2D X-rays}

\def\etal{\emph{et al}\bmvaOneDot}

\begin{document}

\maketitle

\input{sections/abstract}

\input{sections/introduction}

\input{sections/literature_review}

\input{sections/methodology}

\input{sections/experimental_results}


\input{sections/conclusion}

\section*{Acknowledgements}
This research was supported by the German Research Foundation (DFG) and the Munich Center for Machine Learning (MCML). We gratefully acknowledge the computational resources provided by the Leibniz Supercomputing Centre (www.lrz.de).

\bibliography{main}

\input{sections/supplementary}

\end{document}

%% file: sections/abstract.tex
\begin{abstract}
Computed tomography (CT) provides rich 3D anatomical details but is often constrained by high radiation exposure, substantial costs, and limited availability. 
While standard chest X-rays are cost-effective and widely accessible, they only provide 2D projections with limited pathological information. Reconstructing 3D CT volumes from 2D X-rays offers a transformative solution to increase diagnostic accessibility, yet existing methods predominantly rely on synthetic X-ray projections, limiting clinical generalization. 
In this work, we propose AXON, a multi-stage \emph{diffusion}-based framework that reconstructs 3D CT volumes directly from real X-rays with substantially improved fidelity over existing approaches.
AXON adopts a coarse-to-fine reconstruction paradigm that progressively transforms 2D X-ray observations into anatomically consistent 3D volumes. The framework begins with a Brownian Bridge diffusion model that captures global anatomical structures, followed by a ControlNet-guided refinement stage that enhances local intensity details and anatomical realism. To alleviate the depth ambiguities inherent in 2D-to-3D reconstruction, AXON seamlessly incorporates bi-planar X-ray views, enabling more accurate spatial reasoning and structural recovery. A dedicated super-resolution module further increases the spatial resolution of the generated volumes.
Extensive experiments on both public and external datasets demonstrate that AXON consistently surpasses state-of-the-art approaches while maintaining strong generalizability across diverse clinical distributions. Notably, at our highest-resolution bi-planar setting, AXON achieves an $11.9\%$ improvement in PSNR and an $11.0\%$ increase in SSIM over the strongest baseline evaluated at that resolution. In the $128^3$ single-planar setting, it maintains a consistent lead of $7.8\%$ in PSNR on LIDC-IDRI, with even larger margins on the external clinical dataset of $8.0\%$ in PSNR and $16.9\%$ in SSIM. Our code is available at \url{https://github.com/ai-med/AXON/}.

\end{abstract}

%% file: sections/introduction.tex
\section{Introduction}\label{sec:introduction}

\begin{figure}[tb]
  \centering
  \includegraphics[width=0.80\textwidth]{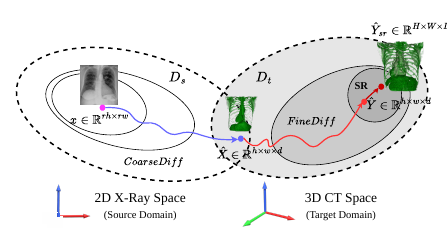}
    \caption{
    AXON reconstructs a high-resolution CT volume from a 2D chest X-ray in three stages: \textit{CoarseDiff} generates an initial CT estimate $\hat{X}$ as an intermediate representation between source and target domains, \textit{FineDiff} refines it into $\hat{Y}$ with enhanced anatomical fidelity, and an optional super-resolution network scales the output by $\gamma = H/h$.
    }
  \label{fig:demo}
  \vspace{-0.8em}
\end{figure}

Computed tomography (CT) is among the most widely used imaging modalities in clinical practice~\cite{OECD2025,sun2010utilisation}. By providing high-resolution, three-dimensional representations of internal anatomy, CT enables accurate diagnosis and treatment planning across a broad range of medical conditions~\cite{liguori_emerging_2015, power2016computed}. Despite its clinical utility, CT imaging comes with notable drawbacks. It exposes patients to substantially higher radiation doses compared to standard radiography~\cite{smith-bindman_radiation_2009}, is more expensive~\cite{hakkaart-van_roijen_costing_2024}, and remains inaccessible in many regions of the world~\cite{hakkaart-van_roijen_costing_2024}. In contrast, conventional X-ray imaging is fast, inexpensive, widely available and portable. However, it captures only two-dimensional projections that provide limited anatomical and pathological information.

Reconstructing a 3D CT volume directly from a single 2D X-ray has the potential to bridge this gap, improving access to detailed volumetric imaging while reducing radiation exposure and clinical costs. However, this task is fundamentally challenging. A single X-ray projection compresses complex 3D anatomy into a 2D image, losing the depth information for sagittal inspection and causing structural overlaps. Variations in acquisition settings, patient positioning, and anatomical variability further complicate the reconstruction problem. Consequently, achieving high-fidelity 3D reconstruction is highly challenging and demands that the model learn the intricate relationships necessary to infer complete volumetric anatomy.

These challenges naturally lead to the use of generative priors, which can capture the distribution of plausible 3D anatomy and reconstruct missing depth information based on limited 2D projections. Diffusion models~\cite{ho2020denoising, song2020denoising} in particular have emerged as a powerful and stable generative framework, outperforming adversarial networks and variational autoencoders in accuracy and robustness. In the medical domain, diffusion-based architectures such as MAISI~\cite{guo_maisi_2025} and Cor2Vox~\cite{bongratz_3d_2025} have shown promising results in generating 3D anatomical structures. Despite this progress, volumetric reconstruction from real X-ray projections remains underexplored. Existing approaches~\cite{ying_x2ct-gan_2019, shen_patient-specific_2019, corona-figueroa_repeat_2024, xu_tomograf_2024, kyung_perspective_2023, liu_volumenerf_2024, tan_xctnet_2022, liu_diffux2ct_2025, bai_xctdiff_2024, jeong_dx2ct_2025, xie_dvg-diffusion_2025} typically evaluate only on synthetic data such as digitally reconstructed radiographs (DRRs)~\cite{galvin1995use}, which differ substantially from clinical X-rays due to idealized physics and noise characteristics. As a result, models trained solely on synthetic data often fail to generalize to real-world applications.

To overcome these limitations, we propose the Advanced X-ray to CT-volume Network (AXON), a coarse-to-fine \emph{diffusion} framework that progressively reconstructs high-resolution CT volumes from 2D chest X-ray images. 
Instead of directly mapping across domains, AXON decomposes the problem into structured intermediate stages that reduce the gap between 2D and 3D domains, and improve anatomical fidelity (Figure~\ref{fig:demo}).
This modular approach enables the seamless integration of robust, pre-trained generative priors and makes the optimization more tractable.
By first generating a plausible volumetric estimate directly from a 2D radiograph and subsequently refining it through conditional CT generation (with an optional super-resolution stage), the proposed AXON framework produces structurally consistent, high-resolution 3D reconstructions while remaining compatible with real clinical radiographs. Our key contributions are:
\begin{itemize}
    \item A novel diffusion-based framework for 2D-to-3D CT generation from sparse single- or bi-planar radiographs, decoupling the reconstruction process to mitigate the severe spatial ambiguity of direct domain mapping.
    \item A hierarchical conditioning and super-resolution strategy that integrates a dedicated conditional 3D diffusion model for localized intensity refinement and a super-resolution network to scale outputs to high resolutions.
    \item 
    Evaluation exclusively on real X-rays from both public and external clinical datasets, extending prior real-world evaluations, which were largely limited to qualitative results or low quantitative accuracy, and demonstrating substantial improvements in PSNR and SSIM over baselines with robust generalizability to real-world clinical data.
\end{itemize}

%% file: sections/literature_review.tex
\section{Related Work}\label{sec::related-work}

\noindent\textbf{Deep Learning Approaches for CT Reconstruction from X-rays} 
span UNet-, GAN-, NeRF-, and diffusion-based paradigms. Shen \etal~\cite{shen_patient-specific_2019} used a ResNet-based encoder for 2D feature extraction, a transformation module to lift features into 3D space, and a generative network for volumetric reconstruction. Building on multi-scale representations, XctNet~\cite{tan_xctnet_2022} performs explicit 2D-to-3D feature fusion. The ``Repeat and Concatenate'' approach~\cite{corona-figueroa_repeat_2024} projects single or multiple X-ray views into 3D space via repetition and concatenation before employing a 3D UNet backbone for reconstruction.
GAN-based methods further enhance perceptual realism through adversarial supervision. Notable examples include X2CT-GAN~\cite{ying_x2ct-gan_2019} for bi-planar 3D reconstruction and the spatial-aware CT synthesis framework by Kyung \etal~\cite{kyung_perspective_2023}.  
NeRF-based approaches explicitly model the radiographic projection process using continuous volumetric representations, as seen in VolumeNeRF~\cite{liu_volumenerf_2024}. These methods enable reconstruction from extremely sparse views via differentiable ray sampling as in TomoGRAF~\cite{xu_tomograf_2024}.
More recently, diffusion models~\cite{ho2020denoising} have emerged as a dominant framework, formulating CT generation as an iterative denoising process conditioned on 2D projections. Existing methods incorporate geometry-aware feature fusion like DIFR3CT~\cite{sun_difr3ct_2024}, tri-plane conditioning mechanisms as DiffuX2CT~\cite{liu_diffux2ct_2025}, and auxiliary view synthesis via latent diffusion as shown with DVG-Diffusion~\cite{xie_dvg-diffusion_2025}. XctDiff~\cite{bai_xctdiff_2024} leverages latent anatomical priors extracted from X-rays to guide the denoising process.
Hybrid pipelines such as X-ray2CTPA~\cite{cahan_x-ray2ctpa_2025} combine generative supervision with specific clinical objectives to improve diagnostic utility.

Unlike supervised conditional models, zero-shot diffusion samplers such as DPS~\cite{chung2023dps}, $\Pi$GDM~\cite{song2023pigdm}, and DDS~\cite{chung2024dds} solve inverse problems by enforcing data consistency through a predefined, differentiable forward operator. Although this operator is perfectly known for synthetic DRRs, real radiographs systematically violate this prerequisite due to scatter, beam hardening, detector effects, and registration inaccuracies, thereby precluding the use of such samplers in our real-X-ray setting.

\noindent\textbf{Synthetic X-rays and Domain Adaptation}. 
DRRs~\cite{sherouse_computation_1990, bahner_digitally_1999} have been largely studied and widely used in clinical applications such as brachytherapy planning~\cite{milickovic_ct_2000}. Due to the lack of large paired CT–X-ray datasets, most recent X-ray-to-CT reconstruction methods rely on DRRs as synthetic projections~\cite{shen_patient-specific_2019, corona-figueroa_repeat_2024, xu_tomograf_2024, liu_volumenerf_2024, liu_diffux2ct_2025, sun_difr3ct_2024, jeong_dx2ct_2025, bai_xctdiff_2024}. Established toolkits such as DeepDRR~\cite{unberath_deepdrr_2018}, Plastimatch~\cite{sharp_plastimatch_2010}, TIGRE~\cite{biguri_tigre_2016}, and DiffDRR~\cite{chen_fast_2023} facilitate realistic DRR generation. However, DRRs differ significantly from real clinical X-rays. Conventional simulation pipelines insufficiently model physical effects such as scatter and beam hardening~\cite{staub_digitally_2013}, while real detectors introduce pixel crosstalk, electronic noise, and vendor-specific preprocessing variations~\cite{unberath_enabling_2019}. Consequently, DRRs lack realistic noise characteristics and anatomical variability~\cite{bai_xctdiff_2024, cahan_x-ray2ctpa_2025}, limiting generalization to clinical data.

To address this domain gap, prior work highlights the discrepancy between DRRs and real radiographs~\cite{sun_difr3ct_2024}. Approaches such as X2CT-GAN~\cite{ying_x2ct-gan_2019} employ CycleGAN-based style transfer~\cite{zhu_unpaired_2017} to map real X-rays into DRR-like images, while others translate DRRs into X-ray style for qualitative evaluation~\cite{kyung_perspective_2023, bai_xctdiff_2024, jeong_dx2ct_2025}. Despite these efforts, bridging the synthetic-to-clinical gap remains an open challenge. Since real X-rays differ substantially from DRRs, assessing performance in clinical settings remains challenging. Reliable evaluation requires testing on real data, both qualitatively and quantitatively. Prior works~\cite{ying_x2ct-gan_2019, kyung_perspective_2023} report qualitative results on real X-rays, but the absence of paired datasets prevents quantitative analysis. 
To date, X-ray2CTPA~\cite{cahan_x-ray2ctpa_2025} is the only method to report quantitative results on real X-ray–CT pairs; however, it merely fine-tunes a model on real X-rays, and its reconstruction accuracy remains limited. This exposes a clear gap between synthetic benchmarking and clinically validated performance. In contrast, AXON is trained jointly on real and synthetic pairs and evaluated exclusively on real X-rays, substantially improving reconstruction fidelity over this fine-tuning-based approach.

%% file: sections/methodology.tex
\section{AXON: Advanced X-ray to CT-volume Network}

\begin{figure}[t]
    \centering
    \includegraphics[width=1.0\textwidth]{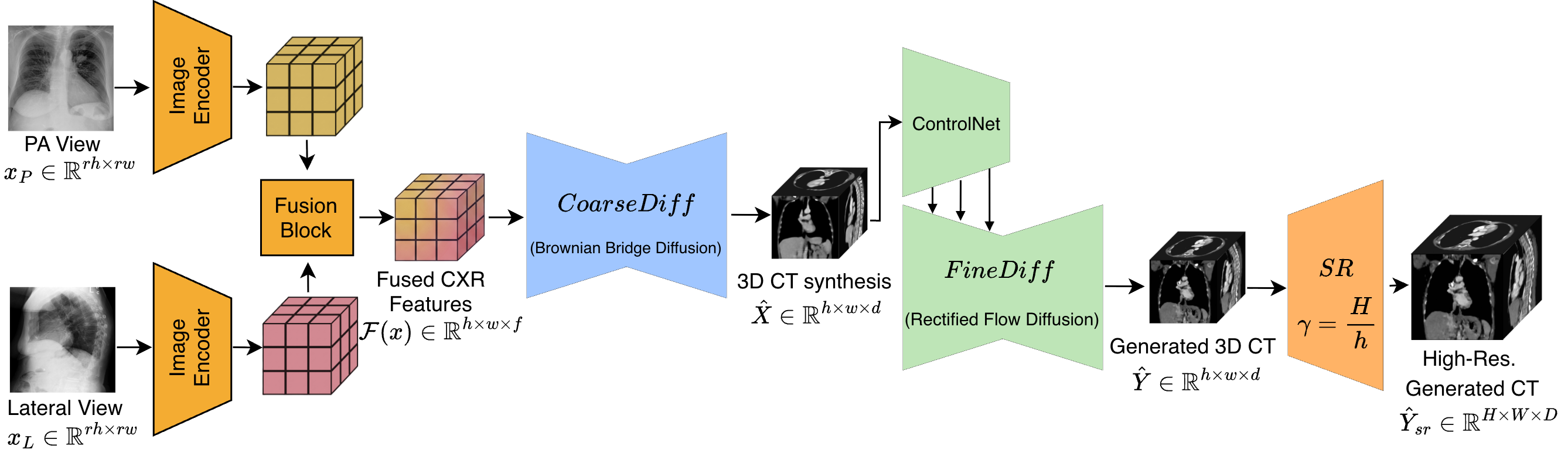}
    \caption{Overview of the proposed cross-dimensional generation framework}
    \label{fig:framework}
    \vspace{-0.3cm}
\end{figure}

Let the source domain be denoted as $\mathcal{D}_s$, consisting of single- or bi-planar images of 2D Chest X-ray (CXR) data $x \in \mathbb{R}^{rh \times rw}$. Its corresponding counterpart in the target domain $\mathcal{D}_t$ is the 3D volumetric CT scan, represented as $Y \in \mathbb{R}^{H \times W \times D}$. $r$ is a scaling factor for cropping the input image from the original chest X-ray. 
The primary objective of the proposed AXON pipeline is to learn a mapping function $G: \mathcal{D}_s \rightarrow \mathcal{D}_t$ that predicts a synthetic CT volume $\hat{Y}$ such that it maximizes the structural and perceptual similarity to the ground-truth target CT $Y$, conditioned on the source input image $x$. The conditioning ensures that the generated internal structures, such as lung parenchyma and vascular networks, are strictly aligned with the radiographic shadows in the source image. Hence, we seek to optimize:
\begin{equation}
    \hat{Y} = \arg\max_{Y_{g} \in \mathcal{D}_t} \mathcal{S}(Y_{g}, Y) \quad \text{subject to} \quad Y_{g} = G(x),
\end{equation}
where $\mathcal{S}$ represents a multi-modal similarity metric. This task is inherently ill-posed and challenging due to the dimensional discrepancy between the 2D planar projections in $\mathcal{D}_s$ and the 3D spatial volumes in $\mathcal{D}_t$, which complicates the generative synthesis process.  
To address the ill-posed nature of 3D CT synthesis from 2D inputs, we propose a two-phase sequential diffusion framework, as shown in Figure~\ref{fig:framework}. In the first phase, \emph{CoarseDiff} (Section~\ref{sec:coarsediff}) explicitly shifts the representation from the 2D domain to the 3D volumetric space. In the second phase, the proposed \emph{FineDiff} (Section~\ref{sec::finediff}) refines the result to resolve the ill-posed reconstruction problem in 3D space. Due to the high computational cost of volumetric diffusion, both stages operate at a relatively low spatial resolution.  
A dedicated 3D super-resolution module is subsequently applied to upsample the generated volumes to higher resolution, as detailed in Section~\ref{sec:sr}. 

\subsection{2D-to-3D Diffusion Model}
\label{sec:coarsediff}

In medical imaging, and particularly in the 2D CXR to 3D CT setting, the domain gap is substantial. The two modalities exhibit inherently heterogeneous characteristics: CXRs represent projective radiographic attenuation maps, whereas CT volumes encode depth-resolved anatomical structures. Our empirical findings indicate that naïvely stacking the 2D input or designing a straightforward 2D-to-3D conditioning branch fails to effectively preserve and transfer the rich structural information of the source domain. To bridge this gap, we introduce \emph{CoarseDiff}, which is based on the Brownian Bridge Diffusion Model (BBDM)~\cite{li2023bbdm}. It is a diffusion framework designed to \textit{explicitly} model transformations between two different domains, rather than generating samples purely from noise. The central idea of this strategy is to first infer a plausible yet coarse volumetric CT representation directly from the input CXR. This intermediate prediction captures the global anatomical structure and overall intensity distribution, while potentially lacking fine-grained structural details. The coarse 3D estimate is subsequently used as a structured condition for the main diffusion model described in Section~\ref{sec::finediff}, which refines it into a high-fidelity CT volume by recovering detailed anatomical features and improving spatial consistency.

As illustrated in Figure~\ref{fig:framework}, we employ an image encoder that extracts high-level semantic features from the input CXR and lifts them into a volumetric representation. The encoder is composed of multiple convolutional layers that progressively downsample the input from $rh \times rw$ pixels to a feature tensor of size $h \times w \times f$, where the channel dimension $f$ is reshaped and interpreted as a depth axis to form a coarse 3D feature volume $\mathcal{F}(x) \in \mathbb{R}^{h \times w\times f}$.
The image encoder is trained jointly with the following BBDM process, forming a unified bridge diffusion model that intrinsically learns to transfer representations across dimensional domains.
Within this co-optimized framework,
BBDM constructs a stochastic bridge between the source sample $\mathcal{F}(x)$ and the ground-truth CT $Y$, making it particularly suitable for cross-domain translation tasks. 
The forward bridging process is expressed as: 
\begin{equation}
x_t = (1 - \alpha_t) x_0 + \alpha_t x_T + \sqrt{\delta_t}\epsilon,\: \:s.t. \:\:\: x_0=Y \:\ \wedge \:x_T=\mathcal{F}(x).
\end{equation}
\noindent Here, $x_t$ denotes the intermediate volume at timestep $t$, $\alpha_t$ controls the interpolation between the clean source and target endpoints, $\delta_t$ represents the variance schedule, and $\epsilon \sim \mathcal{N}(0, I)$ is Gaussian noise. We employ a 3D U-Net $f_\theta$ as the backbone conditioned on $\mathcal{F}(x)$ or equivalently on $x$, with the training objective:
\begin{equation}
\mathcal{L}_{\text{Coarse}} 
= \mathbb{E}_{x, Y, t, \mathcal{N}(0,\mathbf{I})} 
\left[
\left\|
\alpha_t (Y-\mathcal{F}(x)) + \sqrt{\delta_t}\epsilon - f_\theta(x_t, t, \mathcal{F}(x)) 
\right\|_2^2
\right].
\end{equation}

During sampling, we start directly from the conditional input $x_T = \mathcal{F}(x)$, rather than from Gaussian noise $\mathcal{N}(0,\mathbf{I})$ as in standard diffusion models, and iteratively predict $x_{t-1}$ from $x_t$, finally recovering the clean volume $x_0$ through the reverse bridging process. We denote the generated volume as $\hat{X} \in \mathbb{R}^{h \times w \times d}$.
At each timestep $t$, the model predicts the residual term using the diffusion U-Net $f_\theta(x_t, t, \mathcal{F}(x))$, and $x_{t-1}$.  To accelerate inference, we adopt the DDIM strategy~\cite{song2020denoising}, which defines a non-Markovian sampling trajectory while preserving the same marginal distributions as the original Markovian process.

\noindent\textbf{Bi-Planar X-ray Encoding}. The input to our model can also consist of bi-planar X-ray images, both with a resolution of $rh \times rw$ pixels. We employ a shared image encoder to process both views, leveraging weight sharing to project the inputs into a latent feature space with a resolution of $h \times w \times f$. 
To integrate the information from both perspectives, we introduce a \texttt{FusionBlock}. This module processes the concatenated feature volumes through a series of three $3 \times 3 \times 3$ volumetric convolutions, each followed by Group Normalization and ReLU activation. To facilitate gradient flow and preserve spatial features, the block incorporates a residual skip connection via a $1 \times 1 \times 1$ convolution that maps the two input channels to a single output channel. This fusion process results in a unified 3D representation of shape $h \times w \times f$. Subsequent processing stages then follow the standard single-view pipeline.

\subsection{3D-to-3D Diffusion Model}\label{sec::finediff}
While \emph{CoarseDiff} learns to lift lower-dimensional representations into the higher-dimensional space, it struggles to recover fine-grained anatomical details due to the inherently ill-posed nature of 3D synthesis from 2D observations.
To address this, we first train a 3D U-Net unconditionally as a high-capacity generative prior over CT volumes. We adopt a rectified flow formulation~\cite{liu2022flow}, which learns a deterministic transport between a Gaussian source distribution and the CT data distribution along approximately straight trajectories, enabling high-fidelity generation with substantially fewer sampling steps. We define the linear interpolation
\begin{equation}
    y_t = (1-t)\,\epsilon + t\,Y, \qquad t \in [0,1],
\end{equation}
and train a velocity field $v_\theta$ to regress the displacement between the two endpoints:
\begin{equation}
    \mathcal{L}_{Fine} = \mathbb{E}_{Y,\, \epsilon \sim \mathcal{N}(0,\mathbf{I}),\, t \sim \mathcal{U}[0,1]} \left[ \big\| v_\theta(y_t, t, \hat{X}) - (Y - \epsilon) \big\|^2 \right].
\end{equation}
With the 3D U-Net flow model frozen, we incorporate a ControlNet~\cite{zhang2023adding} branch to guide generation using the synthesized intermediate $\hat{X}$ as a structured condition; we call this conditional refinement phase \emph{FineDiff}. During sampling, we initialize $y_0 = \epsilon \sim \mathcal{N}(0,\mathbf{I})$ and integrate the learned ODE $\mathrm{d}y_t/\mathrm{d}t = v_\theta(y_t, t, \hat{X})$ from $t=0$ to $t=1$ with a standard solver, recovering the refined volume $\hat{Y} \in \mathbb{R}^{h \times w \times d}$ in a small number of steps.

\subsection{3D Super-Resolution Model}
\label{sec:sr}
Applying diffusion models to high-resolution data incurs prohibitive computational and memory costs.
This has motivated two common strategies: operating in a compressed latent space, as in Latent Diffusion Models (LDMs)~\cite{Rombach_2022_CVPR}, or training on lower-resolution grids followed by a dedicated super-resolution (SR) upscaling stage~\cite{saharia2022photorealistic}. 
With the increasing prevalence of SR for high-fidelity 4K/8K imagery~\cite{wang2025seedvr,wang2025seedvr2}, we follow this paradigm to address the cubic memory scaling inherent to 3D volumetric generation in both AXON stages. Standard A100 GPUs limit maximum feasible training resolution to $128^3$. Hence, we perform conditioning and generation at this lower scale and subsequently employ an SR module to increase the resolution across all axes. We adapt the U-Net architecture of ESRGAN~\cite{wang2021real} and process the volumetric input slice-wise along the depth dimension. By treating the 3D volume slices as a batch of 2D images, the original 2D network can operate without structural modification. Once all slices are upsampled, the outputs are concatenated to reconstruct the 3D volume. To ensure full volumetric upsampling and maintain 3D global consistency, we apply a 3D Transposed Convolution layer as the final stage. This layer refines the output to a higher voxel resolution. The adapted SR-Net provides a computationally efficient approach to enhance the spatial resolution of generated CT volumes. 
It is worth noting that when the scale ratio $\gamma = \frac{H}{h}$ equals 1, the SR block is not required; however, it becomes necessary for $\gamma \ge 2$.

%% file: sections/experimental_results.tex
\section{Experimental Results}\label{section:experiments}

\subsection{Experimental Setup}\label{section:experimental_setup}

\begin{figure*}[t]
    \centering
    \includegraphics[width=0.86\textwidth]{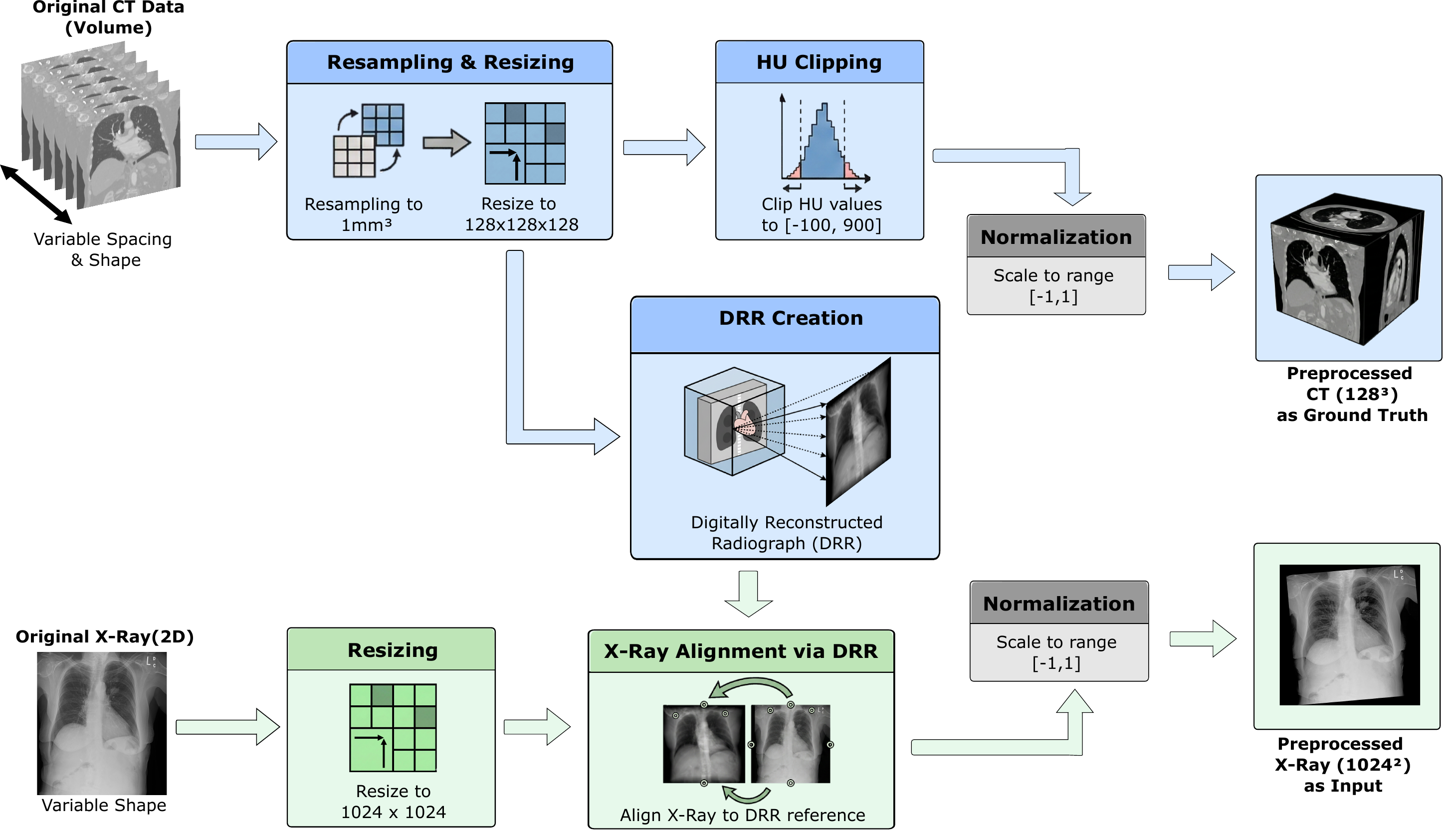}
    \caption{Overview of the data preprocessing workflow}
    \label{fig:preprocessing_pipeline}
\end{figure*}

\paragraph{Dataset and Preprocessing}

We use both the LIDC-IDRI dataset~\cite{armato_iii_lung_2011} and a single-site in-house dataset with paired CT and X-ray images. LIDC-IDRI~\cite{armato_iii_lung_2011} contains a total of 1,018 chest CT scans, where 232 patients also have corresponding real chest X-rays. We only use these paired samples for validation and testing. The resulting LIDC-IDRI split consists of 197 training samples, 10 validation samples, and 25 test samples.
For experiments using bi-planar views as input, we keep the paired samples that include an additional lateral view, resulting in 149 samples for training, 8 for validation, and 16 for testing.
The available X-rays were acquired using either Computed Radiography (CR) or Digital/Direct Capture (DC) systems. CT scans exhibit a fixed in-plane resolution of $512\times512$ pixels, but vary in the number of slices, voxel spacing, and scanner manufacturer. 
We first resample them to achieve a uniform voxel spacing of $1.0\times1.0\times1.0\ \mathrm{mm}^3$, ensuring geometric consistency across scans. 
The volumes were then rescaled to a fixed voxel grid of 
$128\times128\times128$.
To highlight the tissue range relevant for diagnosis, the CT intensities were adjusted to the range $[-100 HU, +900 HU]$, consistent with prior work~\cite{cahan_x-ray2ctpa_2025}. This emphasized lung and vascular structures while saturating high-attenuation structures such as cortical bone. Finally, the intensities were normalized to the range $[-1, 1]$.
The X-ray inputs were standardized to $1024\times1024$.
Since real X-rays are often not spatially aligned with their corresponding CT volumes, a preprocessing step was introduced to improve geometric consistency. As DRRs are aligned with their source CTs, they were used as reference images to guide the registration of the corresponding real X-rays. The complete preprocessing pipeline is shown in Figure~\ref{fig:preprocessing_pipeline} and a paired registered example is shown in supplementary Section~B. A control experiment showing that the reported improvements cannot be attributed to leakage of target geometry through this registration step is provided in supplementary  Section~C.

The same preprocessing pipeline was applied to the in-house dataset, which was entirely unseen during training and used exclusively for an external evaluation.
The acquired CT volumes had a voxel spacing of $1.5 \times 1.5 \times 1.5\ \mathrm{mm}^3$ but no uniform spatial resolution and were acquired using Philips and Siemens scanners. The corresponding X-ray images had resolutions ranging from $1983 \times 1451$ to $3026 \times 3050$ pixels.
From the available X-ray–CT pairs, we selected 220 samples, ensuring adequate patient positioning and full coverage of the chest region (see supplementary Section~A). All of these include an additional lateral view for bi-planar experiments. 
Using an independent and unseen external dataset as a dedicated test set robustly validates the generalizability of the model across diverse acquisition protocols, compensating for the limited sample size of the LIDC-IDRI dataset. This strategic separation enables a larger internal training split.

\paragraph{Synthetic X-ray Generation and Pairing}


To overcome the limited availability of paired real X-rays with CT, the unpaired 786 CT scans from the LIDC-IDRI dataset~\cite{armato_iii_lung_2011} were used to generate synthetic X-ray projections as digitally reconstructed radiography (DRR) using the DeepDRR framework~\cite{unberath_deepdrr_2018}.
Traditional DRRs use ray tracing, which is fast yet unrealistic as it assumes a single material and a mono-energetic spectrum~\cite{russakoff_fast_2005}, meaning that it misses beam hardening and scatter. Alternatively, they use Monte Carlo simulations, which are highly realistic with material-dependent interactions yet slow in processing~\cite{badal_accelerating_2009}.
DeepDRR combines deep learning for 3D material decomposition and 2D scatter estimation with a faster forward projection that accounts for multiple materials, achieving a realistic simulation~\cite{unberath_deepdrr_2018}.
In contrast to prior works that train, evaluate, and test on synthetic data, we integrate synthetic X-rays solely as additional training data, while validation and testing are strictly performed on real X-ray inputs to ensure realistic performance assessment. For all synthetic samples, a single frontal projection was generated to mimic the standard posterior-anterior (PA) X-ray view, consistent with the real images.

\paragraph{Models and Hyperparameters}

The initial AXON-CoarseDiff employs BBDM with $T=1000$ steps, trained for 90 epochs (14 hours) using an $L_2$ loss and an Adam optimizer with a learning rate of $10^{-4}$. The U-Net backbone features 64 base channels and attention layers at $16^3$ and $8^3$ resolutions. The AXON-FineDiff builds on an unconditional 3D U-Net diffusion prior that we train ourselves on the same training split used throughout, with the validation and test volumes held out. It is trained for 320 epochs (78 hours) at a learning rate of $10^{-4}$. It then integrates a ControlNet-based conditioning branch and is trained additionally for 150 epochs (42 hours). Finally, the SR-Net upsamples volumes to $256^3$ via 6 Residual-in-Residual Dense Blocks (RRDB) and 16 initial features, trained for 56 epochs (23.5 hours) with an initial learning rate of $2\times10^{-4}$ and exponential decay. All stages used a batch size of 2 and gradient accumulation to maintain training stability. Since the input CXRs are of higher resolution than the CT data, we set $r = 8$, such that for a $128^3$ generated CT, a $1024^2$ region ($rh\times rw$) is cropped from the original CXR. 
All experiments were implemented on one NVIDIA A100 GPU.
During inference, CoarseDiff uses 10 and FineDiff 100 sampling steps, resulting in a total runtime of approximately 32 seconds per sample on a single A100 GPU.

\paragraph{Baselines}
Our baselines include X2CT-GAN~\cite{ying_x2ct-gan_2019}, a GAN-based approach using synthetic DRR-generated X-rays from CT; XctDiff~\cite{bai_xctdiff_2024}, X-ray2CTPA~\cite{cahan_x-ray2ctpa_2025} and DVG-Diffusion~\cite{xie_dvg-diffusion_2025}, diffusion-based approaches; Repeat and Concatenate~\cite{corona-figueroa_repeat_2024}, a U-Net-based baseline trained on synthetic DRR-generated X-rays. We also include the CoarseDiff-only results of AXON as comparison. 
To ensure a fair comparison, all baselines were directly trained from scratch on the same paired data and with the identical preprocessing pipeline described in Section~\ref{section:experimental_setup}. This represents a substantially harder setting than the one in the original publications, where X2CT-GAN and Repeat and Concatenate were trained and quantitatively evaluated exclusively on synthetic DRRs, and X-ray2CTPA was fine-tuned on real X-rays. DRRs are, by construction, perfectly aligned with their source CT volumes and free of detector noise, scatter, beam hardening, and vendor-specific preprocessing variations. Absolute scores reported here are therefore not directly comparable to those in the respective original papers.

\paragraph{Evaluation}
We evaluate AXON both quantitatively and qualitatively.
A quantitative comparison is conducted between different methods based on the quality metrics, including mean absolute error (MAE), mean squared error (MSE), peak signal-to-noise ratio (PSNR), and structural similarity index measure (SSIM).
Statistical significance between AXON and the baselines is assessed via a two-sided paired Wilcoxon signed-rank test (p < 0.001 across all settings and metrics).
Primary experiments used a resolution of 128 ($\gamma=1$) to ensure fair comparisons with baselines, unless specified otherwise.
In addition to the numerical evaluation, a qualitative analysis provides visual comparisons of reconstructed CT volumes.  
We further show 3D renderings of the generated CT volumes to provide a clearer insight into the 3D consistency of the anatomical structures.

\begin{table}[t]
\centering
\begin{tabular}{llcccc}
\toprule
Data                       & Method                     & MAE (↓)                  & MSE (↓)                  & PSNR (↑)                  & SSIM (↑)                 \\ \midrule
\multirow{6}{*}{\makecell{LIDC-\\IDRI}} & X-ray2CTPA~\cite{cahan_x-ray2ctpa_2025}                & 0.0756{\tiny$\pm$0.0074}          & 0.0181{\tiny$\pm$0.0026}          & 17.47{\tiny$\pm$0.60}              & 0.274{\tiny$\pm$0.046}          \\
                           & R\&C~\cite{corona-figueroa_repeat_2024}    & 0.0575{\tiny$\pm$0.0061}          & 0.0137{\tiny$\pm$0.0022}          & 18.69{\tiny$\pm$0.65}          & 0.361{\tiny$\pm$0.041}          \\
                                
                            & DVG-Diffusion~\cite{xie_dvg-diffusion_2025}    & 0.0496{\tiny$\pm$0.0090}          & 0.0133{\tiny$\pm$0.0036}          & 18.89{\tiny$\pm$1.08}          & 0.504{\tiny$\pm$0.063}          \\
                           
                           & X2CT-GAN~\cite{ying_x2ct-gan_2019}                   & 0.0490{\tiny$\pm$0.0073}          & 0.0118{\tiny$\pm$0.0025}          & 19.37{\tiny$\pm$0.91}              & 0.501{\tiny$\pm$0.059}            \\
                           
                            & XctDiff~\cite{bai_xctdiff_2024}                   & 0.0458{\tiny$\pm$0.0074}          & 0.0110{\tiny$\pm$0.0024}          & 19.68{\tiny$\pm$0.89}              & \textit{0.524{\tiny$\pm$0.067}}            \\
                           \cmidrule{2-6}
                           
                           & AXON-CoarseDiff              & \textit{0.0427{\tiny$\pm$0.0075}}          & \textit{0.0096{\tiny$\pm$0.0024}}          & \textit{20.30{\tiny$\pm$1.02}}          & 0.507{\tiny$\pm$0.060}          \\
                           
                           & AXON & \bftab{0.0391{\tiny$\pm$0.0070}}          & \bftab 0.0078{\tiny$\pm$0.0021} & \bftab 21.21{\tiny$\pm$1.12} & \bftab{0.540{\tiny$\pm$0.053}}          \\ \midrule
\multirow{6}{*}{In-house}  & X-ray2CTPA~\cite{cahan_x-ray2ctpa_2025}        & 0.0584{\tiny$\pm$0.0061}        & 0.0132{\tiny$\pm$0.0024}        & 18.87{\tiny$\pm$0.76}        & 0.377{\tiny$\pm$0.039}        \\
                           & R\&C~\cite{corona-figueroa_repeat_2024}        & 0.0650{\tiny$\pm$0.0051}        & 0.0156{\tiny$\pm$0.0022}        & 18.11{\tiny$\pm$0.59}        & 0.287{\tiny$\pm$0.022}        \\
                           & DVG-Diffusion~\cite{xie_dvg-diffusion_2025}    & 0.0631{\tiny$\pm$0.0055}        & 0.0159{\tiny$\pm$0.0021}        & 18.01{\tiny$\pm$0.56}        & 0.350{\tiny$\pm$0.050}        \\
                           & X2CT-GAN~\cite{ying_x2ct-gan_2019}             & 0.0565{\tiny$\pm$0.0060}        & 0.0131{\tiny$\pm$0.0023}        & 18.90{\tiny$\pm$0.75}        & 0.414{\tiny$\pm$0.043}        \\
                           & XctDiff~\cite{bai_xctdiff_2024}                & 0.0590{\tiny$\pm$0.0077}        & 0.0141{\tiny$\pm$0.0029}        & 18.60{\tiny$\pm$0.86}        & 0.395{\tiny$\pm$0.052}        \\
                           \cmidrule{2-6}
                           & AXON-CoarseDiff                                & \textit{0.0511{\tiny$\pm$0.0065}} & \textit{0.0116{\tiny$\pm$0.0027}} & \textit{19.47{\tiny$\pm$0.99}} & \textit{0.436{\tiny$\pm$0.050}} \\
                           & AXON                                  & \bftab 0.0458{\tiny$\pm$0.0065} & \bftab 0.0094{\tiny$\pm$0.0022} & \bftab 20.41{\tiny$\pm$1.02} & \bftab 0.484{\tiny$\pm$0.046} \\ \bottomrule
\end{tabular}

\caption{Quantitative results on LIDC-IDRI and in-house datasets using single-planar real X-ray inputs.}
\label{tab:comparison_related_work_single}
\end{table}

\subsection{Quantitative Evaluation}\label{sec::quantitative}

Table~\ref{tab:comparison_related_work_single} summarizes quantitative comparison between AXON and baseline approaches. Across both the LIDC-IDRI and in-house datasets, the AXON variants consistently outperform baselines across all metrics.
The AXON-CoarseDiff provides a robust initial reconstruction, while the two-stage AXON pipeline achieves peak quantitative fidelity, reaching a PSNR of 21.21 dB on LIDC-IDRI and 20.41 dB on the in-house dataset.
This represents a substantial $\sim$7.8\% improvement in PSNR and a reduction in MAE of approximately 14.6\% compared to the strongest baseline on LIDC-IDRI, XctDiff.  
On the in-house dataset, where X2CT-GAN is the strongest baseline, AXON improves PSNR by $8.0\%$ and reaches an SSIM of $0.484$, a $16.9\%$ relative increase over X2CT-GAN's $0.414$. These gains highlight the effectiveness of additionally integrating X-ray conditioning to preserve structural and contextual features critical for medical interpretation.
This performance improvement reflects the advantage of the coarse-to-fine multi-scale processing in AXON, which provides detailed volumetric refinement by decoupling the global structural synthesis from local intensity optimization. 
Notably, the results on the in-house dataset were obtained using the model trained exclusively on LIDC-IDRI, using the in-house data solely for testing. The superior performance of AXON on this unseen in-house data confirms its high generalizability and robustness across different datasets and clinical distributions.

\subsection{Qualitative Evaluation}\label{sec::qualitative}

\begin{figure}[t]
\centering
\includegraphics[width=\textwidth]{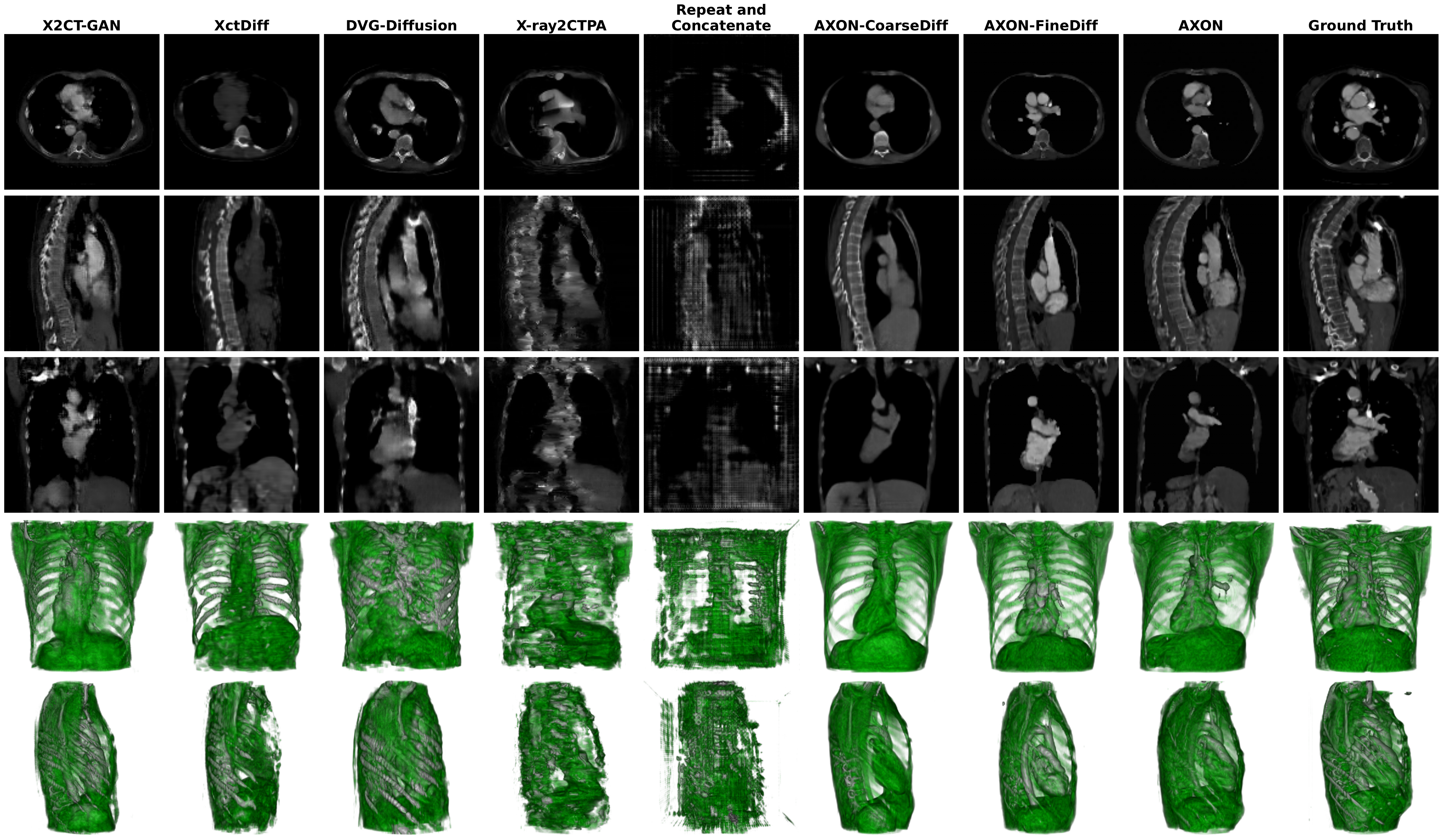}
\caption{Qualitative comparison of 3D CT reconstructions from real X-ray inputs. Axial, sagittal, and coronal mid-slices are shown, followed by 3D volume renderings based on LIDC-IDRI sample 0037.}
\label{fig:qualitative_all}
\end{figure}

\begin{table}[t]
\begin{center}
\setlength{\tabcolsep}{4.2pt}
\begin{tabular}{llcccc}
\toprule
Method & Input   & MAE (↓) & MSE (↓)  & PSNR (↑)  & SSIM (↑)  \\ \midrule
X2CT-GAN~\cite{ying_x2ct-gan_2019}  &   single-planar & 0.0490{\tiny$\pm$0.0073}          & 0.0118{\tiny$\pm$0.0025}          & 19.37{\tiny$\pm$0.91}              & 0.501{\tiny$\pm$0.059}            \\
X2CT-GAN~\cite{ying_x2ct-gan_2019} &  bi-planar   & 0.0435{\tiny$\pm$0.0080}          & 0.0109{\tiny$\pm$0.0027}          & 19.74{\tiny$\pm$0.98}              & 0.517{\tiny$\pm$0.053}            \\

DVG-Diffusion~\cite{xie_dvg-diffusion_2025}  &   single-planar & 0.0496{\tiny$\pm$0.0090}          & 0.0133{\tiny$\pm$0.0036}          & 18.89{\tiny$\pm$1.08}              & 0.504{\tiny$\pm$0.063}            \\
DVG-Diffusion~\cite{xie_dvg-diffusion_2025}  &   bi-planar    & 0.0418{\tiny$\pm$0.0078}          & 0.0106{\tiny$\pm$0.0027}          & 19.86{\tiny$\pm$1.03}              & 0.559{\tiny$\pm$0.056}            \\

 \midrule
 AXON-CoarseDiff              & single-planar & 0.0427{\tiny$\pm$0.0075}          & 0.0096{\tiny$\pm$0.0024}          & 20.30{\tiny$\pm$1.02}          & 0.507{\tiny$\pm$0.060}          \\
AXON-CoarseDiff     &  bi-planar         & \textit{0.0384{\tiny$\pm$0.0073}}          & 0.0082{\tiny$\pm$0.0023}          & 21.02{\tiny$\pm$1.14}          & \textit{0.552{\tiny$\pm$0.063}}          \\
AXON & single-planar &  0.0391{\tiny$\pm$0.0070}          & \textit{0.0078{\tiny$\pm$0.0021}} & \textit{21.21{\tiny$\pm$1.12}} & 0.540{\tiny$\pm$0.053}          \\
AXON &  bi-planar & 
 \bftab
 0.0342{\tiny$\pm$0.0067}          & \bftab 0.0070{\tiny$\pm$0.0018} & \bftab 21.71{\tiny$\pm$1.18} &
\bftab 0.576{\tiny$\pm$0.058} \\ 
\bottomrule        
\end{tabular}
\end{center}
\caption{Results on LIDC-IDRI using \textit{single-} and \textit{bi-planar} real X-ray inputs.}
\label{tab:comparison_related_work_single_bi}
\end{table}

Qualitative analysis in Fig.~\ref{fig:qualitative_all} with more examples in supplementary Section~E.3 reveals structural nuances and anatomical fidelity of the generated CT scans that complement our quantitative findings. Baseline models like Repeat and Concatenate exhibit severe checkerboard artifacts and X-ray2CTPA produces blurred, incoherent structures, indicating that absorbing the imperfections of real X-ray supervision requires the stronger generative prior and decoupled coarse-to-fine formulation introduced in this work.
The proposed AXON models demonstrate superior clarity and anatomical correctness. Specifically, AXON-CoarseDiff provides a structurally coherent foundation, which is further refined by AXON-FineDiff to improve local detail sharpness. The final AXON pipeline generates the most realistic reconstructions, characterized by sharp tissue boundaries and high-fidelity internal structures, such as the branching of the pulmonary vasculature and the distinct borders of the heart chambers. 
Further, as shown in the 3D renderings, AXON effectively preserves global volumetric consistency and complex rib-cage geometry.
While XctDiff, the best-performing baseline on LIDC-IDRI, captures the general lung volume, it fails to maintain sharp structural borders and produces ambiguous anatomy, especially in the heart regions.
In contrast, the AXON pipeline closely approximates the ground truth, generating a continuous and anatomically accurate thoracic appearance.

\subsection{Impact of Bi-planar Projections Input}\label{sec::bi-planar}

The benefit of incorporating additional spatial information is evaluated in Table~\ref{tab:comparison_related_work_single_bi}, which compares single-planar (frontal view) and bi-planar (frontal and lateral views) real X-ray inputs. X-ray2CTPA~\cite{cahan_x-ray2ctpa_2025} was not designed to take more than one X-ray as input. Looking at the single-planar results of R\&C~\cite{corona-figueroa_repeat_2024}, which were already weak, we could not produce reasonable results based on bi-planar X-ray input. Therefore, we chose X2CT-GAN~\cite{ying_x2ct-gan_2019} and DVG-Diffusion~\cite{xie_dvg-diffusion_2025} as representative baselines.

Transitioning from single- to bi-planar inputs consistently enhances all metrics, as the lateral projection helps mitigate the depth ambiguity inherent in 2D-to-3D reconstruction. Adding the second view benefits every method (X2CT-GAN +0.37 dB, DVG-Diffusion +0.97 dB, AXON-CoarseDiff +0.72 dB to reach 21.02 dB). Because AXON already reconstructs at a higher fidelity, with its single-view PSNR alone exceeding the others' bi-planar results, the same absolute gain is correspondingly harder to obtain. This makes its 0.72 dB improvement no less substantial than the larger increments reported at lower PSNR levels. The full AXON pipeline achieves the highest overall fidelity with bi-planar inputs, reaching a peak SSIM of 0.576 and an MAE of 0.0342. Notably, the AXON single-planar results still substantially outperform the bi-planar X2CT-GAN and DVG-Diffusion, underscoring the effectiveness of AXON in high-fidelity CT reconstruction and information integration. Bi-planar results on the external in-house dataset are reported in supplementary Section~E.1.

\subsection{Evaluation of the Super-resolution Network}\label{sec::sr}

Table~\ref{tab:comparison_srnet} reports the quantitative evaluation of CT reconstruction with a higher resolution at $256^3$ ($\gamma=2$).
Transitioning to a higher spatial resolution naturally increases sensitivity to voxel-wise error metrics, as evidenced by X2CT-GAN's PSNR drop from 19.37 dB to 18.79 dB.
Our proposed super-resolution network (SR-Net) integrated in the AXON preserves its high generation fidelity at high resolution.
Among the single-planar configurations, when trained exclusively on ground truth CT scans (w/o Syn), integrating SR-Net achieves a peak SSIM of 0.570, indicating that learning directly from real data pairs optimizes structural preservation. Further, incorporating synthesized outputs generated by AXON during training (w/ Syn) yields the highest intensity accuracy with a PSNR of 20.69 dB, likely because the model learns to compensate for detailed artifacts generated in the preceding stages. 
The most substantial performance gains are observed when integrating SR-Net in the bi-planar configuration, which achieves a peak PSNR of 22.03 dB and an SSIM of 0.628. This performance not only surpasses all single-planar high-resolution variants but also exceeds the results of the $128^3$ bi-planar diffusion stage, confirming that the SR-Net effectively leverages the enhanced spatial constraints of dual-view inputs to resolve anatomical details. 
Ultimately, these results validate that the bi-planar coarse-to-fine AXON framework, coupled with SR-Net refinement, provides a robust and scalable solution for high-resolution 3D CT reconstruction.

\begin{table}[t]
\begin{center}
\begin{tabular}{llcccc}
\toprule
Method & Res. & {MAE (↓)} & {MSE (↓)} &
{PSNR (↑)} & {SSIM (↑)} \\
\midrule
X2CT-GAN\cite{ying_x2ct-gan_2019} {\tiny(\textit{single-planar})}  &    128   & 0.0490{\tiny$\pm$0.0073}          & 0.0118{\tiny$\pm$0.0025}          & 19.37{\tiny$\pm$0.91}              & 0.501{\tiny$\pm$0.059}            \\
X2CT-GAN\cite{ying_x2ct-gan_2019} {\tiny(\textit{bi-planar})}  &    128   & 0.0435{\tiny$\pm$0.0080}          & 0.0109{\tiny$\pm$0.0027}          & 19.74{\tiny$\pm$0.98}              & 0.517{\tiny$\pm$0.053}            \\
AXON {\tiny(\textit{single-planar})} & 128 & 0.0391{\tiny$\pm$0.0070}          & 0.0078{\tiny$\pm$0.0021} & 21.21{\tiny$\pm$1.12} & 0.540{\tiny$\pm$0.053}          \\
AXON {\tiny(\textit{bi-planar})} &  128 & 
 \bftab
 0.0342{\tiny$\pm$0.0067}          & \bftab 0.0070{\tiny$\pm$0.0018} & \bftab 21.71{\tiny$\pm$1.18} &
\bftab 0.576{\tiny$\pm$0.058} \\ 

\midrule
X2CT-GAN\cite{ying_x2ct-gan_2019} {\tiny(\textit{single-planar})} & 256 & 0.0499\tiny$\pm$0.0075 & 0.0135\tiny$\pm$0.0031 & 18.79\tiny$\pm$0.93 & 0.494\tiny$\pm$0.044 \\
X2CT-GAN\cite{ying_x2ct-gan_2019} {\tiny(\textit{bi-planar})} & 256 & 0.0423\tiny$\pm$0.0076 & 0.0110\tiny$\pm$0.0027 & 19.69\tiny$\pm$0.96 & 0.566\tiny$\pm$0.050 \\
AXON$_{\gamma=2}$ {\tiny(\textit{single-planar}, w/ Syn)} & 256 
& 0.0436\tiny$\pm$0.0071 &  \textit{0.0089\tiny$\pm$0.0027} &  \textit{20.69\tiny$\pm$1.26} & 0.491\tiny$\pm$0.069 \\
AXON$_{\gamma=2}$ {\tiny(\textit{single-planar}, w/o Syn)} & 256 &  \textit{0.0414\tiny$\pm$0.0075} & 0.0093\tiny$\pm$0.0027 & 20.50\tiny$\pm$1.21 &  \textit{0.570\tiny$\pm$0.056} \\
AXON$_{\gamma=2}$ {\tiny(\textit{bi-planar}, w/o Syn)} &  256 & \bftab 0.0336\tiny$\pm$0.0057 & \bftab 0.0064\tiny$\pm$0.0014 & \bftab 22.03\tiny$\pm$0.88 & \bftab 0.628\tiny$\pm$0.056   \\
\bottomrule
\end{tabular}
\end{center}
\caption{Results of integrating different SR-Net training strategies for a high output resolution (Res.) of $256$ (AXON$_{\gamma=2}$) for real X-ray inputs from LIDC-IDRI.}
\label{tab:comparison_srnet}
\end{table}

\subsection{Ablation Study}\label{sec::ablation}

Table~\ref{tab:ablation} demonstrates the ablation study on the validation set to evaluate the influence of architectural components, loss formulations, and training configurations in different stages of AXON. Additional ablations are provided in supplementary Section D.
For the initial CoarseDiff stage, we compared the standard $L_2$ loss with an additional edge-loss term ($\mathcal{L}_{2} + \mathcal{L}_\text{edge}$). The inclusion of edge loss resulted in a negligible improvement in PSNR (+0.08 dB) and no significant change in SSIM. Given its limited impact and increased computational overhead, we adopted the $L_2$-only formulation.
For the following FineDiff stage, we evaluated two variants utilizing a mix of real X-rays and synthetic DRRs for training, including: (i) Output Clamping (+clamping): Restricting model outputs to the interval $[-1, 1]$ significantly enhanced SSIM to 0.737 but at a substantial cost to intensity accuracy, with PSNR dropping to 19.91 dB. This suggests that while clamping suppresses artifacts, it limits the dynamic range necessary for precise Hounsfield Unit (HU) mapping. (ii) Image Encoder: Replacing the standard repeat-and-concatenate  strategy (R\&C enc.) with the same image encoder (image enc.) from AXON-CoarseDiff stage resulted in lower performance across all metrics compared to the simpler R\&C approach. 
The unclamped R\&C configuration provided the best balance between voxel-wise accuracy and structural generation with a PSNR of 20.86 dB, and was thus selected for the final pipeline.
The full AXON pipeline with integration of both stages was tested under various conditioning strategies. 
Results indicate that applying the same Brownian bridge diffusion process in the FineDiff stage as in the CoarseDiff stage yields suboptimal performance.
This validates AXON's transition to a conditional 3D diffusion model in the FineDiff stage to be more effective at generating fine-grained, high-fidelity anatomical details.
Removing the additional global encoder (global enc.) layers leads to a higher PSNR of 21.76 dB, suggesting that direct conditioning through the diffusion process provides sufficient guidance without redundant encoding. Furthermore, training on a combined dataset of real X-rays and synthetic DRRs outperformed training only on the real ones (w/o DRRs), likely due to the increased data diversity, which enhances model robustness.


%% file: sections/conclusion.tex
\section{Conclusion}

In this work, we introduced a novel multi-stage diffusion-based framework designed to address the ill-posed challenge of reconstructing 3D CT volumes from sparse 2D X-ray projections. 
In contrast to prior work that trains and quantitatively evaluates almost exclusively on synthetic DRRs, AXON is validated entirely on paired real-world X-ray/CT data, establishing a new baseline for clinically realistic X-ray-to-CT translation. By leveraging the generative priors of diffusion models through a coarse-to-fine strategy, we successfully decoupled global structural synthesis from local intensity refinement. 
Our experimental results on both the public LIDC-IDRI and external unseen in-house datasets demonstrated that AXON substantially outperformed existing state-of-the-art baselines. 
A key advantage of our method is its superior structural integrity and perceptual fidelity. Unlike competing methods, which are prone to stochastic artifacts and boundary blurring, AXON preserved sharp tissue boundaries and complex anatomical geometries such as the thoracic cage and pulmonary vasculature. Furthermore, the model demonstrated exceptional generalizability, maintaining high performance on unseen in-house clinical data despite being trained exclusively on the public dataset. The integration of the super-resolution network further validated the framework's scalability, providing a stable upsampling path to high-resolution generation. 

\begin{table}[t]
\centering

\begin{tabular}{llcccc}
\toprule
CoarseDiff &
FineDiff &
{MAE ($\downarrow$)} & {MSE ($\downarrow$)} &
{PSNR ($\uparrow$)} & {SSIM ($\uparrow$)} \\
\midrule

\multicolumn{6}{l}{\textit{(i) AXON-CoarseDiff only}} \\
\midrule
$\mathcal{L}_2$        & —
  & 0.0427\tiny$\pm$0.0075
  & 0.0096\tiny$\pm$0.0024
  & 20.30\tiny$\pm$1.02
  & \bftab 0.507\tiny$\pm$0.060 \\
$\mathcal{L}_2$ + $\mathcal{L}_\text{edge}$     & —
  & \bftab 0.0417\tiny$\pm$0.0070
  & \bftab 0.0094\tiny$\pm$0.0024
  & \bftab 20.38\tiny$\pm$1.05
  & \bftab 0.507\tiny$\pm$0.059 \\

\midrule
\multicolumn{6}{l}{\textit{(ii) AXON-FineDiff only}} \\
\midrule
—  & + clamping 
  & 0.0428\tiny$\pm$0.0068
  & 0.0104\tiny$\pm$0.0023
  & 19.91\tiny$\pm$0.94
  & \bftab 0.737\tiny$\pm$0.070 \\
—  & image enc.
  & 0.0394\tiny$\pm$0.0067
  & 0.0092\tiny$\pm$0.0025
  & 20.49\tiny$\pm$1.11
  & 0.556\tiny$\pm$0.052 \\
—  & R\&C enc.
  & \bftab 0.0392\tiny$\pm$0.0038
  & \bftab 0.0083\tiny$\pm$0.0010
  & \bftab 20.86\tiny$\pm$0.53
  & 0.516\tiny$\pm$0.059 \\

\midrule
\multicolumn{6}{l}{\textit{(iii) AXON}} \\
\midrule
$\mathcal{L}_2$  & BBDM
  & 0.0410\tiny$\pm$0.0055
  & 0.0094\tiny$\pm$0.0016
  & 20.32\tiny$\pm$0.77
  & 0.515\tiny$\pm$0.056 \\
 $\mathcal{L}_2$  & + global enc.
  & 0.0420\tiny$\pm$0.0068
  & 0.0088\tiny$\pm$0.0022
  & 20.68\tiny$\pm$1.09
  & 0.517\tiny$\pm$0.049 \\
 $\mathcal{L}_2$  & w/o DRRs
  & \textit{0.0390\tiny$\pm$0.0079}
  & 0.0080\tiny$\pm$0.0025
  & 21.18\tiny$\pm$1.28
  & \bftab 0.543\tiny$\pm$0.054 \\
$\mathcal{L}_2$ + $\mathcal{L}_\text{edge}$  & + global enc.
  & 0.0419\tiny$\pm$0.0068
  & 0.0088\tiny$\pm$0.0022
  & 20.70\tiny$\pm$1.11
  & 0.517\tiny$\pm$0.049 \\
$\mathcal{L}_2$ + $\mathcal{L}_\text{edge}$  & w/o DRRs
  & 0.0393\tiny$\pm$0.0077
  & 0.0081\tiny$\pm$0.0025
  & 21.11\tiny$\pm$1.28
  & 0.541\tiny$\pm$0.053 \\
$\mathcal{L}_2$ + $\mathcal{L}_\text{edge}$  &  w/ DRRs
  & 0.0392\tiny$\pm$0.0072
  & \textit{0.0079\tiny$\pm$0.0022}
  & \textit{21.19\tiny$\pm$1.20}
  & 0.538\tiny$\pm$0.054 \\
 $\mathcal{L}_2$  &  w/ DRRs
  & \bftab 0.0372\tiny$\pm$0.0039
  & \bftab 0.0068\tiny$\pm$0.0010
  & \bftab 21.76\tiny$\pm$0.73
  & \textit{0.540\tiny$\pm$0.048} \\
\bottomrule
\end{tabular}
\caption{Ablation study across architectural design choices and input configurations.}
\label{tab:ablation}
\end{table}

\noindent\textbf{Limitations and Future Work.} 
Reconstruction is performed at a maximum resolution of $256^3$, which exceeds the $128^3$ setting common in prior X-ray-to-CT work but remains below native clinical CT resolution; scaling further is constrained by the cubic memory growth of 3D diffusion. In addition, our evaluation relies on image-quality metrics complemented by qualitative inspection; anatomy-specific downstream tasks would provide stronger evidence of clinical utility and are left for future investigation. 
Throughout this work, ``high fidelity'' refers to relative improvement over
existing baselines rather than absolute agreement with the ground truth.
Finally, while we observed no overt hallucinations, the one-to-many nature of X-ray-to-CT translation implies that structures in ambiguous regions may be plausible yet non-veridical, a risk to be assessed per downstream task.

%% file: sections/supplementary.tex
\clearpage

\appendix

\section*{Supplementary Materials}
\renewcommand{\figurename}{Appendix Fig.}
\renewcommand{\tablename}{Appendix Table}
\counterwithin{figure}{section}
\counterwithin{table}{section}
\setcounter{figure}{0}
\setcounter{table}{0}

\section{In-house Dataset Collection}\label{sec:inhouse_data}
The in-house dataset was collected retrospectively at TUM Klinikum Rechts der Isar. We included patients for whom a chest X-ray and a CT scan were acquired within 2 days of each other, with adequate patient positioning and full coverage of the chest region, yielding 220 paired samples. All data were pseudonymized prior to processing. Dataset characteristics and preprocessing are described in Section~4.1 of the main paper.

\section{X-ray Registration Visualization}

\begin{figure}[h]
\centering
\includegraphics[width=\textwidth]{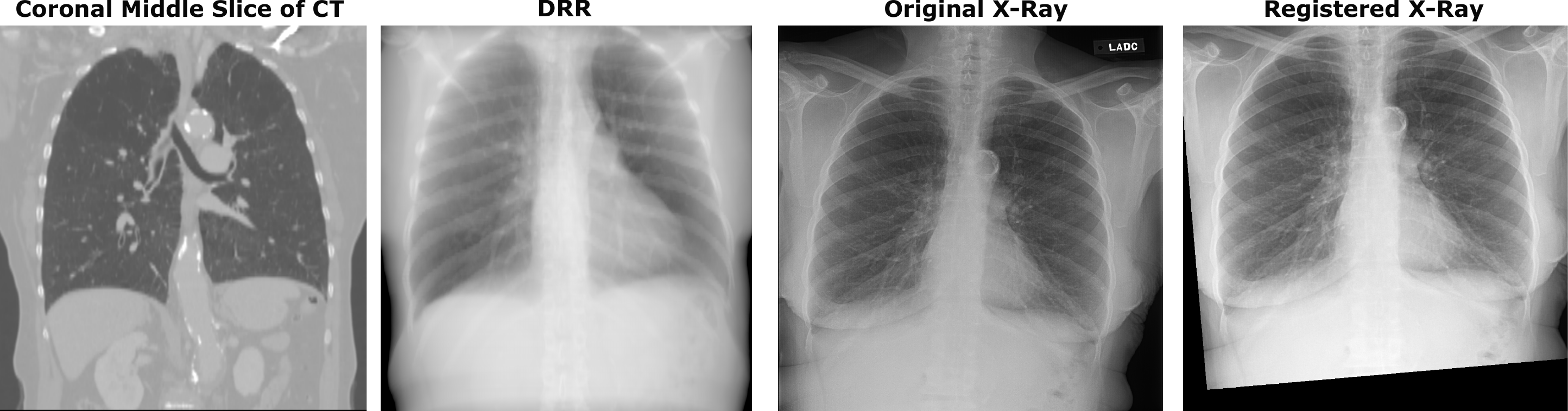}
\caption{Visualization of the registration process. The figure displays the middle coronal CT slice and its corresponding DRR, alongside the original and registered X-ray views.}
\label{fig:x_ray_registration}
\vspace{-0.2cm}
\end{figure}

As explained in Section~4.1 of the main paper, we utilize synthetically generated DRRs to improve the alignment between the X-ray views and their corresponding CT scans. This is because DRRs are inherently aligned with the CT volumes from which they are projected. See Figure~\ref{fig:x_ray_registration} for an example registration of an X-ray based on the DRR of its corresponding CT scan. 

\section{Controlling for Target Geometry Leakage}\label{sec:supp_leakage}
It is important to consider whether the registration of real X-rays to DRRs derived from their paired CTs (Section~4.1 of the main paper) introduces target geometry into the input. To control for this, the evaluation was repeated using an alignment independent of the target: each input X-ray was registered to a fixed template, and the generated volume was subsequently rigidly registered to the ground truth for metric computation only. As shown in Table~C.1, performance under this target-independent protocol decreases moderately in SSIM while PSNR changes only marginally ($-0.49$ dB single-planar, $+0.13$ dB bi-planar), since the rigid post-hoc alignment restores intensity agreement but not fine structural overlap. AXON still exceeds all baselines evaluated under the standard protocol, so the reported improvements cannot be attributed to leakage of target geometry through the registration step.

\begin{table}[h]
\setlength{\tabcolsep}{5pt}
\begin{center}
\begin{tabular}{lcccc}
\toprule
Method & MAE ($\downarrow$) & MSE ($\downarrow$) & PSNR ($\uparrow$) & SSIM ($\uparrow$) \\
\midrule
\multicolumn{5}{l}{\textit{DRR-based Registration}} \\
\midrule
AXON {\tiny(\textit{single-planar})} & 0.0458{\tiny$\pm$0.0065} & 0.0094{\tiny$\pm$0.0022} & 20.41{\tiny$\pm$1.02} & 0.484{\tiny$\pm$0.046} \\
AXON {\tiny(\textit{bi-planar})}     & 0.0420{\tiny$\pm$0.0064} & 0.0079{\tiny$\pm$0.0021} & 21.17{\tiny$\pm$1.11} & 0.508{\tiny$\pm$0.047} \\
\midrule
\multicolumn{5}{l}{\textit{Template Registration}} \\
\midrule
AXON {\tiny(\textit{single-planar})} & 0.0495{\tiny$\pm$0.0084} & 0.0105{\tiny$\pm$0.0028} & 19.92{\tiny$\pm$1.15} & 0.445{\tiny$\pm$0.067} \\
AXON {\tiny(\textit{bi-planar})}     & 0.0435{\tiny$\pm$0.0074} & 0.0077{\tiny$\pm$0.0022} & 21.30{\tiny$\pm$1.22} & 0.464{\tiny$\pm$0.062} \\
\bottomrule
\end{tabular}
\end{center}
\caption{Comparison of DRR-based and target-independent template registration
on the in-house dataset ($128^3$).}
\label{tab:template_registration}
\end{table}

\section{Additional Ablation Study}
\counterwithin{figure}{section}
\counterwithin{table}{section}
To optimize feature representation for bi-planar X-ray views, we evaluated three Image Encoder architectures (Table~\ref{tab:ablation_imageencoder}). We compared two Late-Fusion approaches that process X-rays (resolution of $1024 \times 1024$) into two $128^3$ shaped feature volumes, which are then merged via a \textit{FusionBlock} as described in Section 3.1 of the main paper. The first approach, used in our final pipeline, employs a single convolutional network with shared weights for both views, which processes them into two $128^3$ feature volumes. The second approach uses the same architecture but with two separate pipelines for each view. Finally, we tested an Early-Fusion approach. In this setup, the X-ray views are first repeated and concatenated according to their respective dimensions, then first fused and downsampled afterwards into the $128^3$ feature volume. To address GPU memory constraints that prevent processing a $1024^3$ volume, we first downscale the X-rays to $256 \times 256$ before the repetition and concatenation steps.
Comparing the two Late-Fusion approaches, the model with shared weights performs best. This suggests that forcing the encoder to learn a unified feature space for both views is more effective than using two dedicated pipelines. 
A larger performance gap is observed when employing the Late-Fusion approaches compared to the Early-Fusion approach. This difference may be attributed in part to the initial processing requirements of the Early-Fusion architecture. Due to GPU memory constraints, the original $1024 \times 1024$ X-ray images must be downscaled to $256 \times 256$ prior to repetition and concatenation. The downscaling process likely discards fine-grained structural details before the network processes the images.

\begin{table}[h]
\setlength{\tabcolsep}{5pt}
\begin{center}
\begin{tabular}{lcccc}
\toprule
Image Encoder  & MAE (↓) & MSE (↓)  & PSNR (↑)  & SSIM (↑)\\ 
\midrule
Early-Fusion &  
 0.0431{\tiny$\pm$0.0054}          &  0.0106{\tiny$\pm$0.0017} &  19.79{\tiny$\pm$0.73} &
 0.488{\tiny$\pm$0.049} \\
Late-Fusion (no shared weights) &  
 0.0428{\tiny$\pm$0.0050}          &  0.0093{\tiny$\pm$0.0016} &  20.36{\tiny$\pm$0.75} &
 0.503{\tiny$\pm$0.054} \\
Late-Fusion (shared weights)  & \bftab 0.0412{\tiny$\pm$0.0062}          & \bftab 0.0090{\tiny$\pm$0.0016}          & \bftab 20.51{\tiny$\pm$0.76}          & \bftab 0.526{\tiny$\pm$0.070}         \\
\bottomrule
\end{tabular}
\end{center}
\caption{Ablation Study for different Image Encoder types for AXON-CoarseDiff using bi-planar X-ray inputs on validation dataset.}
\label{tab:ablation_imageencoder}
\end{table}

\section{Additional Experiments}
\counterwithin{figure}{section}
\counterwithin{table}{section}

\subsection{Bi-planar Results on the In-house Dataset}

Table~\ref{tab:inhouse_biplanar} extends the bi-planar comparison of Section~4.4 of the main paper to the external in-house dataset. Consistent with the LIDC-IDRI results, all methods benefit from the additional lateral view, and AXON achieves the highest fidelity across all metrics. Notably, AXON-CoarseDiff alone already exceeds both bi-planar baselines, and the full pipeline adds a further gain of more than 1\,dB in PSNR, mirroring the coarse-to-fine behavior observed on LIDC-IDRI.

\begin{table}[h]
\setlength{\tabcolsep}{5pt}
\begin{center}
\begin{tabular}{lcccc}
\toprule
Method & MAE ($\downarrow$) & MSE ($\downarrow$) & PSNR ($\uparrow$) & SSIM ($\uparrow$) \\
\midrule
DVG-Diffusion & 0.0566{\tiny$\pm$0.0050} & 0.0136{\tiny$\pm$0.0019} & 18.71{\tiny$\pm$0.58} & 0.411{\tiny$\pm$0.052} \\
X2CT-GAN  & 0.0551{\tiny$\pm$0.0050} & 0.0128{\tiny$\pm$0.0021} & 18.98{\tiny$\pm$0.72} & 0.409{\tiny$\pm$0.037} \\
AXON-CoarseDiff          & 0.0495{\tiny$\pm$0.0057} & 0.0099{\tiny$\pm$0.0021} & 20.12{\tiny$\pm$0.89} & 0.447{\tiny$\pm$0.053} \\
AXON                     & 0.0420{\tiny$\pm$0.0064} & 0.0079{\tiny$\pm$0.0021} & 21.17{\tiny$\pm$1.11} & 0.508{\tiny$\pm$0.047} \\
\bottomrule
\end{tabular}
\end{center}
\caption{Results on the in-house dataset using bi-planar real X-ray inputs
($128^3$).}
\label{tab:inhouse_biplanar}
\end{table}

\subsection{Contribution of Individual Pipeline Stages}

To justify the design choice of a combined multi-stage pipeline using bi-planar X-ray images rather than a single-stage diffusion approach, we evaluated each component individually. The Image Encoder detailed in Section 3.1 of the main paper was used across all three experiments. As demonstrated by the results in Table~\ref{tab:appendix_additional_experiments}, the AXON pipeline outperforms both the AXON-CoarseDiff and AXON-FineDiff stages trained in isolation. The AXON pipeline yields a $3.3\%$ increase in PSNR and a $4.3\%$ increase in SSIM over AXON-CoarseDiff. Compared to AXON-FineDiff, the combined approach achieves improvements of $5.2\%$ in PSNR and $10.6\%$ in SSIM.

\begin{table}[h]
\setlength{\tabcolsep}{5pt}
\begin{center}
\begin{tabular}{lcccc}
\toprule
Method  & MAE (↓) & MSE (↓)  & PSNR (↑)  & SSIM (↑)\\ 
\midrule
AXON-CoarseDiff only  & 0.0384{\tiny$\pm$0.0073}          & 0.0082{\tiny$\pm$0.0023}          & 21.02{\tiny$\pm$1.14}          & 0.552{\tiny$\pm$0.063}         \\
AXON-FineDiff only &  

 0.0407{\tiny$\pm$0.0089}          & 0.0091{\tiny$\pm$0.0031} &  20.63{\tiny$\pm$1.41} &
 0.521{\tiny$\pm$0.065} \\
AXON & 
 \bftab
 0.0342{\tiny$\pm$0.0067}          & \bftab 0.0070{\tiny$\pm$0.0018} & \bftab 21.71{\tiny$\pm$1.18} &
\bftab 0.576{\tiny$\pm$0.058} \\ 
\bottomrule
\end{tabular}

\end{center}
\caption{Test Results on LIDC-IDRI using bi-planar X-ray inputs (output resolution $128^3$).}\label{tab:appendix_additional_experiments}
\end{table}

\clearpage

\subsection{Additional Qualitative Results}

\begin{figure}[h]
\centering
\includegraphics[width=\textwidth]{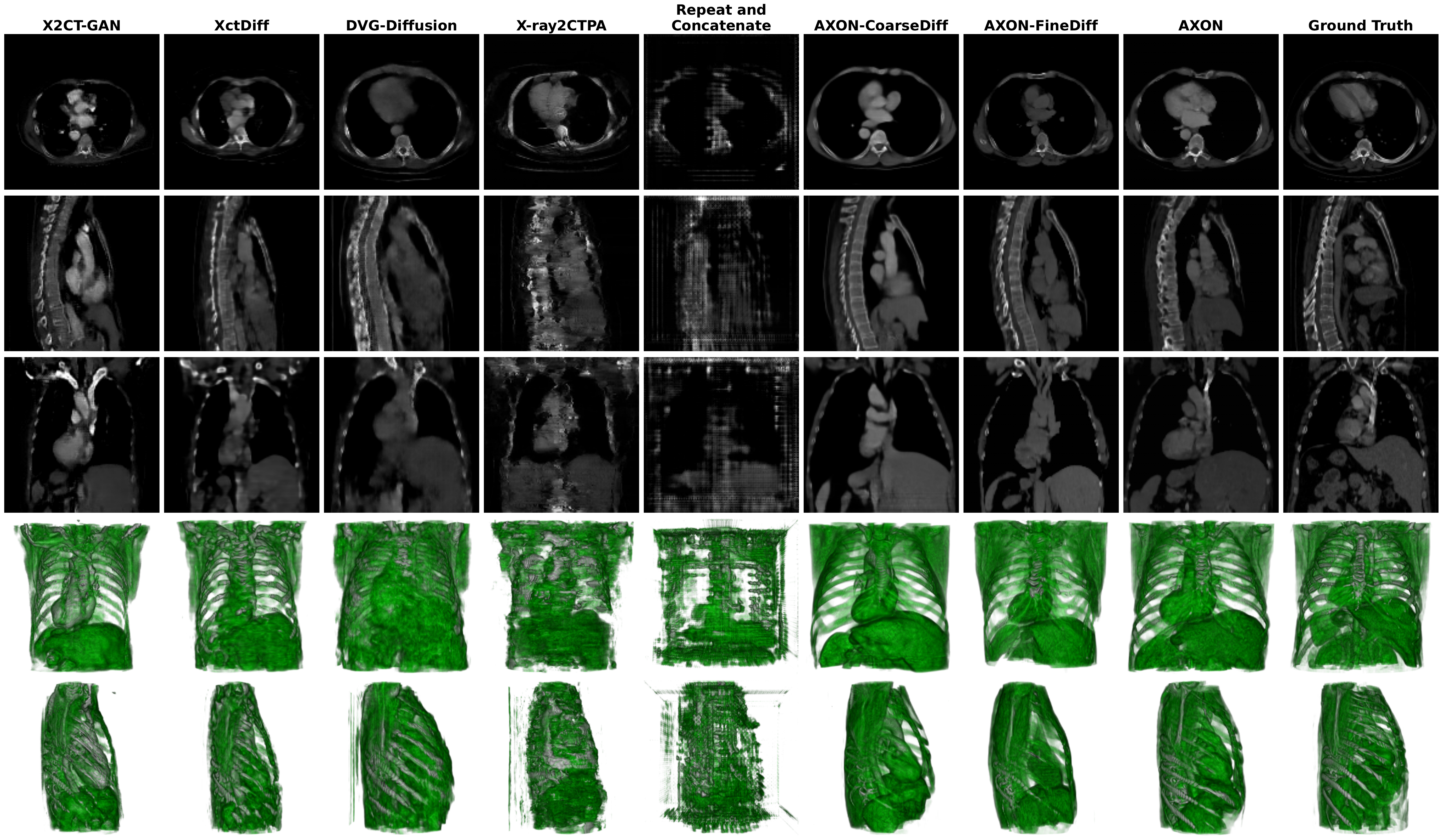}
\caption{Additional Qualitative comparison of 3D CT reconstructions from real X-ray inputs. Axial, sagittal, and coronal mid-slices are shown, followed by 3D volume renderings based on LIDC-IDRI sample 0031.}
\label{fig:qualitative_LIDC-IDRI-0031}
\vspace{-0.2cm}
\end{figure}

\begin{figure}[h]
\centering
\includegraphics[width=\textwidth]{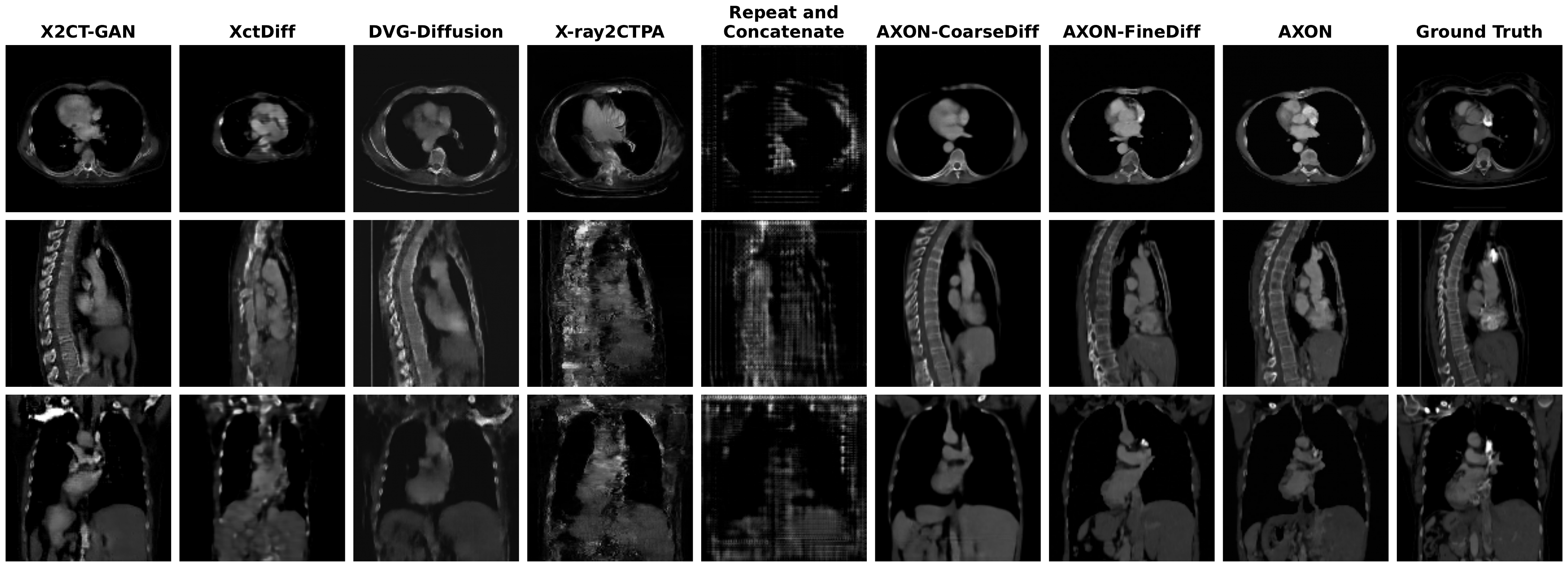}
\caption{Additional Qualitative comparison of 3D CT reconstructions from real X-ray inputs. Axial, sagittal, and coronal mid-slices are shown based on LIDC-IDRI Sample 0193.}
\label{fig:qualitative_LIDC-IDRI-0193}
\vspace{-0.2cm}
\end{figure}

\begin{figure}[h]
\centering
\includegraphics[width=\textwidth]{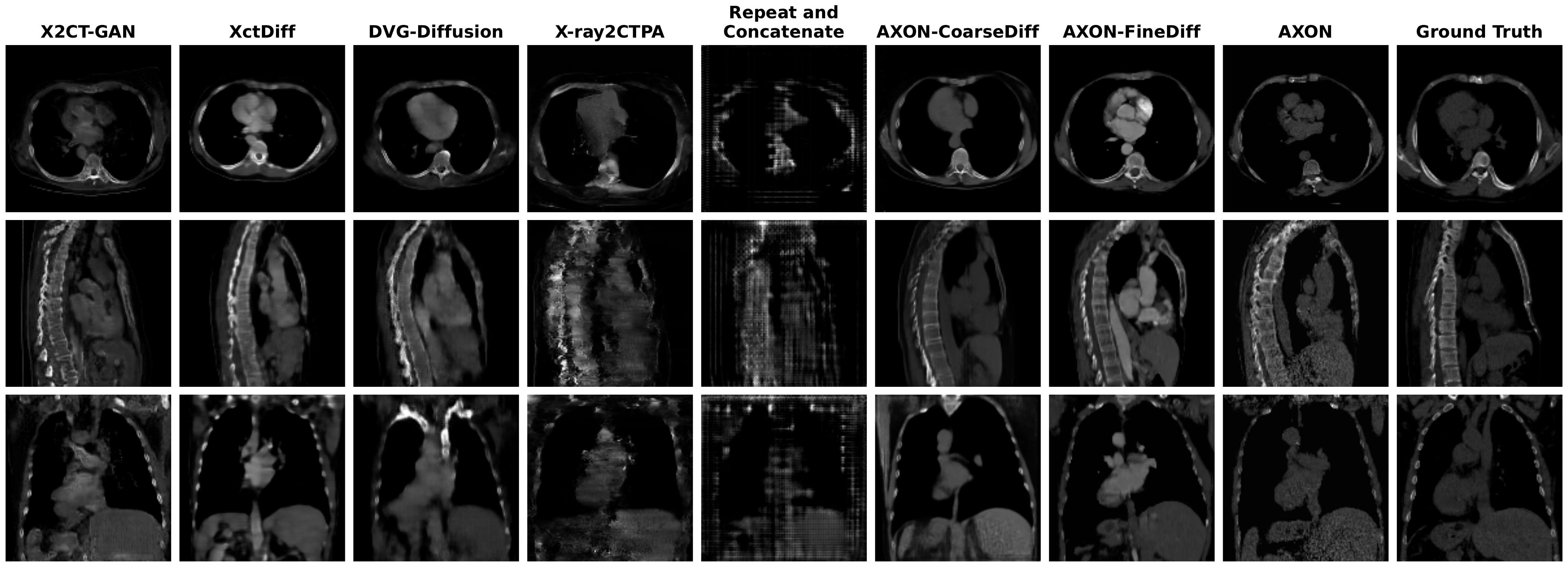}
\caption{Additional Qualitative comparison of 3D CT reconstructions from real X-ray inputs. Axial, sagittal, and coronal mid-slices are shown based on an in-house data sample.}
\label{fig:qualitative_uJ4u4Qw59Pc}
\vspace{-0.2cm}
\end{figure}

\begin{figure}[h]
\centering
\includegraphics[width=\textwidth]{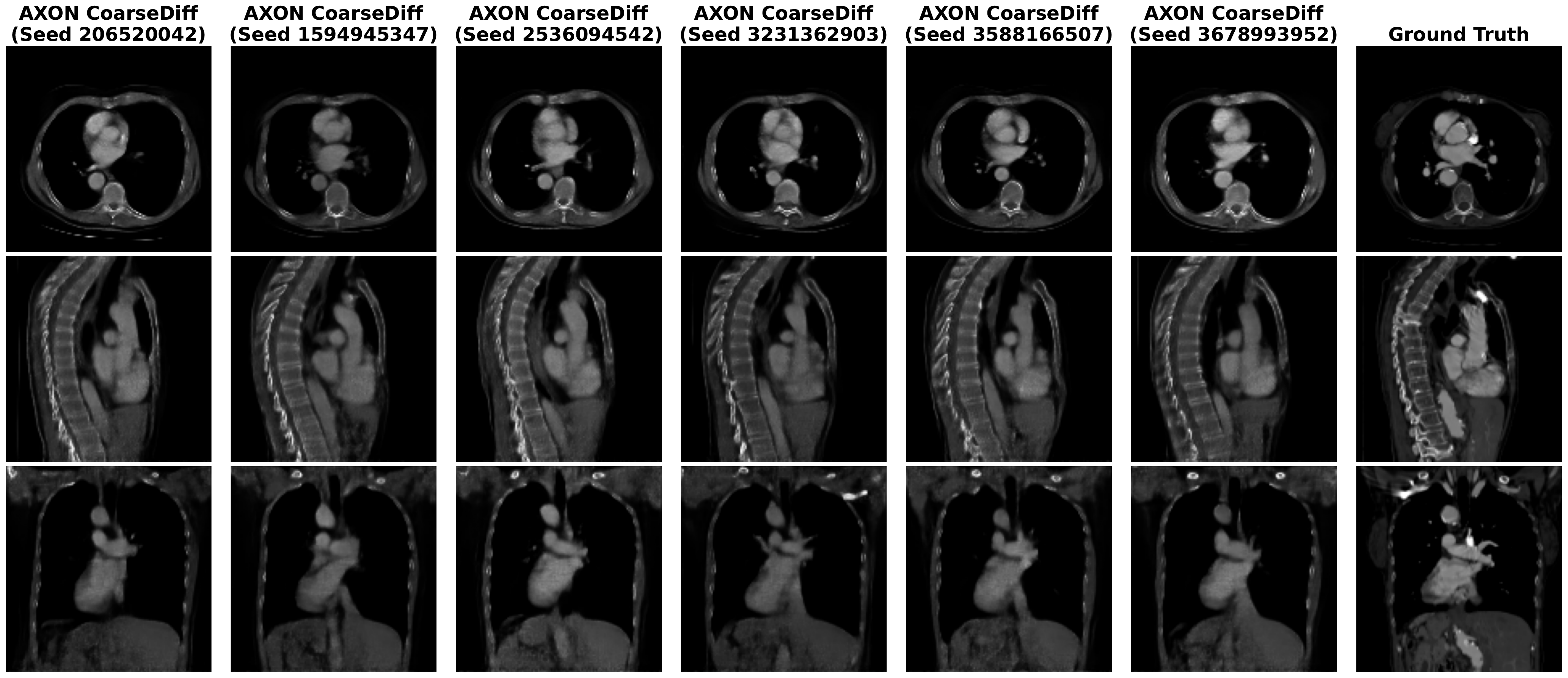}
\caption{Additional Qualitative comparison of 3D CT reconstructions from bi-planar X-ray inputs using AXON-CoarseDiff. Axial, sagittal, and coronal mid-slices are shown based on LIDC-IDRI Sample 0037 using different initialization seeds.}
\label{fig:qualitative_seed_comparison_0037}
\vspace{-0.2cm}
\end{figure}

\begin{figure}[h]
\centering
\includegraphics[width=\textwidth]{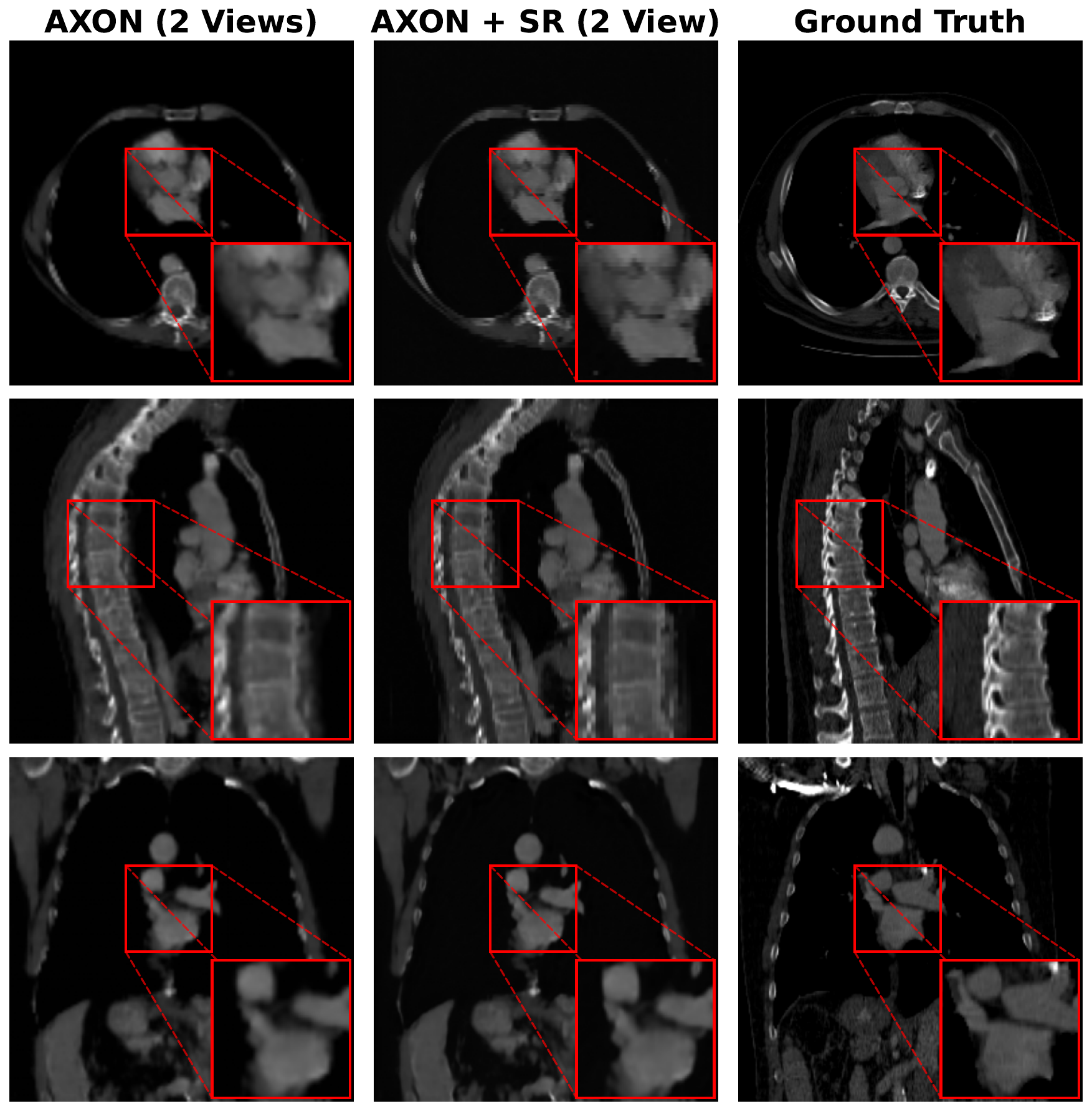}
\caption{Additional Qualitative comparison of 3D CT reconstructions from real bi-planar X-ray inputs. Axial, sagittal, and coronal mid-slices are shown based on LIDC-IDRI Sample 0231 using AXON Pipeline with output resolution $128^3$ ($\gamma=1$) and AXON + SR with output resolution $256^3$ ($\gamma=2$).}
\label{fig:comparison_lr_hr_LIDC-IDRI-0231}
\vspace{-0.2cm}
\end{figure}